\documentclass[3p,times,procedia]{elsarticle}
\usepackage{ecrc}

\usepackage{color}
\usepackage{graphicx}
\usepackage{booktabs}
\usepackage{subfigure}
\usepackage{listings}
\usepackage{multirow}
\usepackage{mathtools}
\usepackage{amsmath}
\usepackage{algorithm}
\usepackage{algorithmic}
\usepackage{gensymb}
\volume{00}

\firstpage{1}

\journalname{}

\runauth{Etemad et al.}

\jid{trpro}

\usepackage{amssymb}
\biboptions{authoryear}

\usepackage[figuresright]{rotating}
\usepackage[bookmarks=false]{hyperref}
    \hypersetup{colorlinks,
      linkcolor=blue,
      citecolor=blue,
      urlcolor=blue}

\begin{document}
\begin{frontmatter}

%% Title, authors and addresses

%% use the tnoteref command within \title for footnotes;
%% use the tnotetext command for the associated footnote;
%% use the fnref command within \author or \address for footnotes;
%% use the fntext command for the associated footnote;
%% use the corref command within \author for corresponding author footnotes;
%% use the cortext command for the associated footnote;
%% use the ead command for the email address,
%% and the form \ead[url] for the home page:
%%
%% \title{Title\tnoteref{label1}}
%% \tnotetext[label1]{}
%% \author{Name\corref{cor1}\fnref{label2}}
%% \ead{email address}
%% \ead[url]{home page}
%% \fntext[label2]{}
%% \cortext[cor1]{}
%% \address{Address\fnref{label3}}
%% \fntext[label3]{}

\dochead{}%

\title{On feature selection and evaluation of transportation mode prediction strategies}

%% use optional labels to link authors explicitly to addresses:
%% \author[label1,label2]{<author name>}
%% \address[label1]{<address>}
%% \address[label2]{<address>}

\author[a]{Mohammad Etemad\corref{cor1}} 
\author[a]{Am\'ilcar Soares J\'unior }
\author[a,b]{Stan Matwin}

\address[a]{Institute for Big Data Analytics, Dalhousie University, Halifax, NS, Canada}
\address[b]{Institute for Computer Science, Polish Academy of Sciences, Warsaw and Postcode, Poland}

\begin{abstract}

Transportation modes prediction is a fundamental task for decision making in smart cities and traffic management systems. 
Traffic policies designed based on trajectory mining can save money and time for authorities and the public. 
It may reduce the fuel consumption and commute time and moreover, may provide more pleasant moments for residents and tourists. 
Since the number of features that may be used to predict a user transportation mode can be substantial, finding a subset of features that maximizes a performance measure is worth investigating. 
In this work, we explore wrapper  and information retrieval methods to find the best subset of trajectory features.
After finding the best classifier and the best feature subset, our results were compared with two related papers that applied deep learning methods and the results showed that our framework achieved better performance. 
Furthermore, two types of cross-validation approaches were investigated, and the performance results show that the random cross-validation method provides optimistic results.

\end{abstract}

\begin{keyword}
%Type your keywords here, separated by semicolons ; 
Feature engineering; Trajectory classification; Trajectory Mining;Transportation Modes Prediction
%% keywords here, in the form: keyword \sep keyword

%% PACS codes here, in the form: \PACS code \sep code

%% MSC codes here, in the form: \MSC code \sep code
%% or \MSC[2008] code \sep code (2000 is the default)

\end{keyword}
\cortext[cor1]{Corresponding author. Tel.: +1-902-418-9420 }
\end{frontmatter}

%\correspondingauthor[*]{Corresponding author. Tel.: +0-000-000-0000 ; fax: +0-000-000-0000.}
\email{etemad@dal.ca}

%%
%% Start line numbering here if you want
%%
% \linenumbers

%% main text

%\enlargethispage{-7mm}
\section{Introduction}
%\label{Introduction}

Trajectory mining is a very hot topic since positioning devices are now used to track people, vehicles, vessels, natural phenomena, and animals. 
It has applications including but not limited to transportation mode detection \citep{zheng2010understanding,endo2016deep,dabiri2018inferring,xiao2017identifying, etemad2018predicting}, fishing detection \citep{de2016improving}, tourism \citep{feng2017poi2vec}, and animal behaviour analysis \citep{fossette2010spatio}. 
There are also a number of topics in this field that need to be investigated further such as high performance trajectory classification methods \citep{endo2016deep,dabiri2018inferring,zheng2010understanding,xiao2017identifying,liu2017end}, accurate trajectory segmentation methods \citep{zheng2008understanding,soares2015grasp,grasp-semts2018},  trajectory similarity and clustering \citep{kang2009similarity}, dealing with trajectory uncertainty \citep{hwang2018segmenting}, active learning \cite{soares-2017}, and semantic trajectories \citep{parent2013semantic}. 
These topics are highly correlated and solving one of them requires to some extent exploring the more than one. 
For example, to perform a trajectory classification, it is necessary to deal with noise and segmentation directly and the other topics mentioned above indirectly.

As one of the trajectory mining applications, transportation modes prediction is a fundamental task for decision making in smart cities and traffic management systems. 
Traffic policies designed based on trajectory mining can save money and time for authorities and the public. 
It may reduce the fuel consumption and commute time and moreover, may provide more pleasant moments for residents and tourists. 
Since a trajectory is a collection of geo-locations captured through the time, extracting features that show the behavior of a trajectory is of prime importance. 
The number of features that can be generated for trajectory data is significant. 
However, some of these features are more important than others for the transportation mode prediction task. 
Selecting the best subset of features not only save the processing time but also may increase the performance of the learning algorithm. 
The features selection problem and the trajectory classification task were selected as the focus of this research. 
The contributions of this work are listed below. 
\begin{itemize}
  \item Using two feature selection approaches, we investigated the best subset of features for transportation modes prediction.
  \item After finding the best classifier and the best subset of features, we compare our results with the works of \cite{dabiri2018inferring} and \cite{endo2016deep}. The results showed that our approach performed better than the others from literature.
  \item Finally, we investigate the differences between two methods of cross-validation used by the literature of transportation mode prediction. The results show that the random cross-validation method suggests optimistic results in comparison to user-oriented cross-validation.
\end{itemize}
 
The rest of this work is structured as follows. 
The related works are reviewed in section \ref{sec:PreviousWork}. 
The basic concepts, definitions and the framework we designed are provided in section \ref{sec:not}. 
%The Geolife dataset is described in section \ref{sec:GeoLifeDataset}. 
%Section \ref{sec:UncertaintyofData} talks about handling noise and harnessing the uncertainty of data. 
%The framework used in this work is detailed in section \ref{sec:model}. 
We provide our experimental results in section \ref{sec:experiments}. 
Finally, the conclusions and future works are shown in section \ref{sec:conclusion}.

\section{Related works}
\label{sec:PreviousWork}

Feature engineering is an essential part of building a learning algorithm. Some of the algorithms extract features using representation learning methods; On the other hand, some studies select a subset from the handcrafted features. 
Both methods have advantages such as learning faster, less storage space, performance improvement of learning, and generalized models building \citep{li2017feature}. 
These two methods are different from two perspectives.
First, extracting features generates new features while selecting features chooses a subset of existing features.
Second, selecting features constructs more readable and interpretable models than extracting features \citep{li2017feature}. 
This work focuses on the feature selection task.

Feature selection methods can be categorized into three general groups: filter methods, wrapper methods, and embedded methods \citep{fs_guyon2003introduction}.  
Filter methods are independent of the learning algorithm. 
They select features based on the nature of data regardless of the learning algorithm \citep{li2017feature}. 
On the other hand, wrapper methods are based on a kind of search, such as sequential, best first, or branch and bound, to find the best subset that gives the highest score on a selected learning algorithm \citep{li2017feature}. 
The embedded methods apply both filter and wrapper \citep{li2017feature} such as decision tree.
Feature selection methods can be grouped based on the type of data as well. 
The feature selection methods that use the assumption of i.i.d.(Independent and identically distributed) are conventional feature selection methods \citep{li2017feature} such as \cite{he2005laplacian} and \cite{zhao2007spectral}.
They are not designed to handle heterogeneous or auto-correlated data. 
Some feature selection methods have been introduced to handle heterogeneous data and stream data that most of them working on graph structure such as  \cite{gu2011towards}. Conventional feature selection methods are categorized in four groups: similarity-based methods like \cite{he2005laplacian}, Information theoretical methods like \cite{peng2005feature}, sparse learning methods such as \cite{li2012unsupervised}, and statistical based methods like \cite{liu1995chi2}.
Similarity-based feature selection approaches are independent of the learning algorithm, and most of them cannot handle feature redundancy or correlation between features. Likewise, statistical methods like chi-square cannot handle feature redundancy, and they need some discretization strategies. The statistical methods are also not effective in high dimensional space. Since our data is not sparse and sparse learning methods need to overcome the complexity of optimization methods, they were not a candidate for experiments.
On the other hand, information retrieval methods can handle both feature relevance and redundancy. Furthermore, selected features can be generalized for learning tasks. Information gain, which is the core of Information theoretical methods, assumes that samples are independently and identically distributed. 
Finally, the wrapper method only sees the score of the learning algorithm and try to maximize the score of the learning algorithm.
Therefore, we perform two experiments using a wrapper method and a information theoretical method.

The most common evaluation metric reported in the related works is the accuracy of the models. 
Therefore, we use accuracy metric to compare our work with theirs. 
Since the data was imbalanced, we reported the F score as well. 
Despite the fact that most of the related work applied the accuracy metric, it is calculated using different methods including random cross-validation, cross-validation with dividing users, cross-validation with mix users and simple division of the training and test set without cross-validation. 
The latter is a weak method that is used only in \cite{zhu2018transportation}. 
The random cross-validation or the conventional cross-validation was applied in \cite{xiao2017identifying}, \cite{liu2017end} , and \cite{dabiri2018inferring}. 
\cite{zheng2010understanding} mixed the training and test set according to users so that 70\% of trajectories of a user goes to the training set and the rest goes to test set. 
Only \cite{endo2016deep} performed the cross-validation by dividing users between the training and test set. 
Because trajectory data is a kind of data with spatiotemporal dimensions and the possibility of having users in the same semantic hierarchical structure such as students, worker, visitors, and teachers, the conventional cross-validation method could provide optimistic results as studied in \cite{roberts2017cross}. 
Similar to previous studies, we choose the Geolife dataset and transportation modes detection task. 
However, we investigate the effects of different cross-validation techniques.

\section{Preliminaries}
\label{sec:not}

\subsection{Notations and definitions}

A \emph{trajectory point}, $l_i \in L$, so that $l_i=(x_i,y_i,t_i)$, 
%is defined in notation \ref{not:1}, 
where $x_i$ is longitude varies from 0$^{\circ}$ to $\pm 180^{\circ}$, $y_i$ is latitude varies from 0$^{\circ}$ to $\pm 90^{\circ}$, and $t_i$ ($t_i < t_{i+1}$) is the capturing time of the moving object and $L$ is the set of all trajectory points. 
A trajectory point can be assigned by some features that describe different attributes of the moving object with a specific time-stamp and location. 
The time-stamp and location are two dimensions that make trajectory point  \emph{spatio-temporal} data with two important properties: (i) \emph{auto-correlation} and (ii) \emph{heterogeneity} \cite{STDM2017}. These features makes the conventional cross validation invalid \cite{roberts2017cross}.

A \emph{raw trajectory}, or simply a trajectory, is a sequence of trajectory points captured through time. $\tau=(l_i,l_{i+1},..,l_n), l_j \in L, i \leq n$.
A \emph{sub-trajectory} is one of the consecutive sub-sequences of a raw trajectory generated by splitting the raw trajectory into two or more sub-trajectories. 
For example, if we have one split point, $k$, and $\tau_1$ is a raw trajectory then $s_1=(l_i,l_{i+1},...,l_{k})$ and $s_2=(l_{k+1},l_{k+2},...,l_n)$ are two sub trajectories generated by $\tau_1$. 
The process of generating sub trajectories from a raw trajectory is called \emph{segmentation}.
We used a daily segmentation of raw trajectories and then segmented the data utilizing the transportation modes annotations to partition the data. This approach is also used in  \cite{dabiri2018inferring} and \cite{endo2016deep}. 
The assumption that the transportation modes are available for test set segmentation is invalid since we are going to predict them by our model; 
However, we need to prepare a controlled environment similar to \cite{dabiri2018inferring} and \cite{endo2016deep} to study the feature selection.% performance of the transportation modes prediction.

A \emph{point feature} is a measured value $F_p$, assigned to each trajectory points of a sub trajectory $S$. 
$Fˆp=(f_i,f_{i+1},..,f_n)$ shows the feature $F_p$ for sub trajectory $S$. 
For example, speed can be a point feature since we can calculate the speed of a moving object for each trajectory point. 
Since we need two trajectory points to calculate speed, we assume the speed of the first trajectory point is equal to the speed of the second trajectory point.

A \emph{trajectory feature} is a measured value $F_t$, assigned to a sub trajectory, $S$. 
$F_t= \frac{\Sigma f_k}{n}$
shows the feature $F_t$ for sub trajectory $S$. 
For example, the speed mean can be a trajectory feature since we can calculate the speed mean of a moving object for a sub trajectory. 

The $F_t^p$ is the notation for all trajectory features that generated using point feature $p$. For example, $F_t^{speed}$ represents all the trajectory features derived from $speed$ point feature. Moreover, $F_{mean}^{speed}$ denotes the mean of the trajectory features derived from the $speed$ point feature.

\subsection{The framework}
\label{sec:model}

\begin{figure}[ht]
\centering
\includegraphics[scale=0.38]{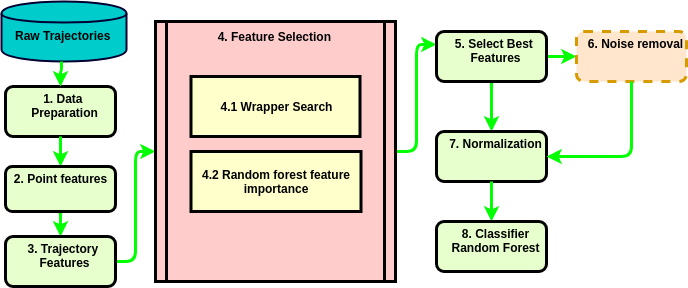}
\caption{The steps of the applied framework to predict transportation modes.}
\label{fig:model}
\end{figure}

In this section, the sequence of steps of the framework with eight steps are explained (Figure \ref{fig:model}).

The first step groups the trajectory points by \textit{user id}, \textit{day} and \textit{transportation modes} to create sub trajectories (segmentation). 
Sub trajectories with less than ten trajectory points were discarded to avoid generating low-quality trajectories. 

Point features including speed, acceleration, bearing, jerk, bearing rate, and the rate of the bearing rate were generated in step two.
The features speed, acceleration, and bearing were first introduced in \cite{zheng2008understanding}, and jerk was proposed in \cite{dabiri2018inferring}. 
The very first point feature that we generated is duration. 
This is the time difference between two trajectory points. 
This feature gives us essential information including some of the segmentation position points, loss signal points, and is useful in calculating point features such as speed, and acceleration.
The distance was calculated using the haversine formula.
Having duration and distance as two point features, we calculate speed, acceleration and jerk using Equation
$S_{i}=\frac{Distance_{i}}{Duration_{i}}$,
$A_{i+1}=\frac{(S_{i+1}-S_{i})}{\Delta t}$, and
$J_{i+1}=\frac{(A_{i+1}-A_{i})}{\Delta t}$
%\ref{eq:speed},\ref{eq:acc}, and \ref{eq:jerk},
respectively.
A function to calculate the bearing ($B$) between two consecutive points was also implemented. 
Two new features were introduced in \cite{etemad2018predicting}, named bearing rate, and the rate of the bearing rate. 
Applying 
$B_{rate(i+1)}=\frac{(B_{i+1}-B_{i})}{\Delta t}$,
%equation \ref{eq:5}, 
we computed the bearing rate. 
$B_i$ and $B_{i+1}$ are the bearing point feature values in points $i$ and $i+1$.
$\Delta t$ is the time difference.% \cite{etemad2018predicting}. 
The rate of the bearing rate point feature is computed using 
$Br_{rate(i+1)} =\frac{(B_{rate(i+1)}-B_{rate(i)})}{\Delta t}$.
Since extensive calculations are done with trajectory points, it was necessary an efficient way to calculate all these equations for each trajectory. 
Therefore, the code was written in a vectorized manner in Python programming language which is faster than other online available versions.

After calculating the point features for each trajectory, the trajectory features were extracted in step three. 
Trajectory features are divided into two different types including global trajectory features and local trajectory features.
Global features, like the Minimum, Maximum, Mean, Median, and Standard Deviation, summarize information about the whole trajectory and local trajectory features, like percentiles (e.g., 10, 25, 50, 75, and 90), describe a behavior related to part of a trajectory. 
The local trajectory features extracted in this work were the percentiles of every point feature.
Five different global trajectory features were used in the models tested in this work. 
In summary, we compute 70 trajectory features (i.e., 10 statistical measures including five global and five local features calculated for 7 point features) for each transportation mode sample. 
In Step 4, two feature selection approaches were performed, wrapper search and information retrieval feature importance. According to the best accuracy results for cross-validation, a subset of top 20 features was selected in step 5.
The code implementation of all these steps is available at \hyperlink{trajlib}{https://github.com/metemaad/TrajLib}.

In step 6, the framework deals with noise in the data optionally. This means that we ran the experiments with and without this step.
Finally, we normalized the features (step 7) using the Min-Max normalization method, since this method preserves the relationship between the values to transform features to the same range and improves the quality of the classification process \citep{han2011data}. 

\section{Experiments}
\label{ch:experiments}
\label{sec:experiments}

In this section, we detail the four experiments performed in this work to investigate the different aspects of our framework. 
In this work, we used the GeoLife dataset \citep{zheng2008understanding}.
This dataset has 5,504,363 GPS records collected by 69 users, and is labeled with eleven transportation modes: taxi (4.41\%); car (9.40\%); train (10.19\%); subway (5.68\%); walk (29.35\%); airplane (0.16\%); boat (0.06\%); bike (17.34\%); run (0.03\%); motorcycle (0.006\%); and bus (23.33\%).
Two primary sources of uncertainty of the Geolife dataset are device and human error. % which are reviewed in GPS records captured by a device always have some inaccuracy. 
This inaccuracy can be categorized in two major groups, \emph{systematic errors} and \emph{random errors} \citep{jun2006smoothing}. 
The \textit{systematic error} occurs when the recording device cannot find enough satellites to provide precise data. 
The \textit{random error} can happen because of atmospheric and ionospheric effects. 
Furthermore, the data annotation process has been done after each tracking as \cite{zheng2008understanding} explained in the Geolife dataset documentation.
As humans, we are all subject to fail in providing precise information; it is possible that some users forget to annotate the trajectory when they switch from one transportation mode to another. 
For example, the changes in the speed pattern (changes in the size of marker) might be a representation of human error.

We assume the \emph{bayes error} is the minimum possible error and human error is near to the \emph{bayes error} \cite{ng2016nuts}.
Avoidable bias is defined as the difference between the training error and the human error.  Achieving the performance near to the human performance in each task is the primary objective of the research. The recent advancements in deep learning lead to achieving some performance level even more than the performance of doing the task by human because of using large samples and scrutinizing the data to fine clean it. 
However, ``we cannot do better than \emph{bayes error} unless we are overfitting". \cite{ng2016nuts}.
Having noise in GPS data and human error suggest the idea that the avoidable bias is not equal to zero. 
This ground truth was our base to include research results in our related work or exclude it.

The user-oriented cross-validation and the random forest classifier were used for evaluation of transportation modes used in \cite{endo2016deep}.
The wrapper method implemented to search the best subset of our 70 features.
The information theoretical feature importance methods were used to select the best subset of our 70 features for the transportation modes prediction task. 
The third experiment is a comparison between \cite{endo2016deep} and our implementation. The user-oriented cross-validation, the top 20 best features, and random forest were applied to compare our work with \cite{endo2016deep}.
The random cross-validation on the top 20 features was applied to classify transportation modes used in \cite{dabiri2018inferring} using a random forest classifier.

\subsection{Classifier selection}

\begin{figure}[h]
\centering
 \begin{tabular}{ll}
\includegraphics[scale=0.31]{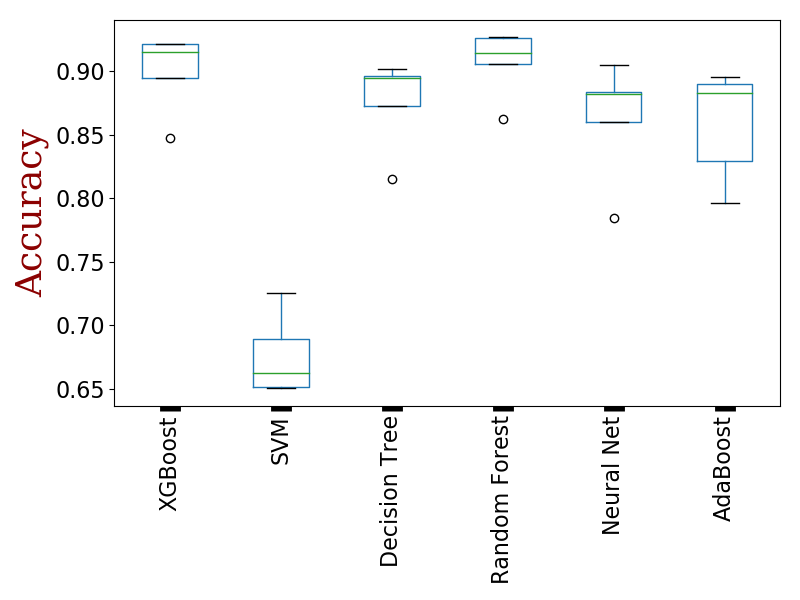}
 &
 \includegraphics[scale=0.31]{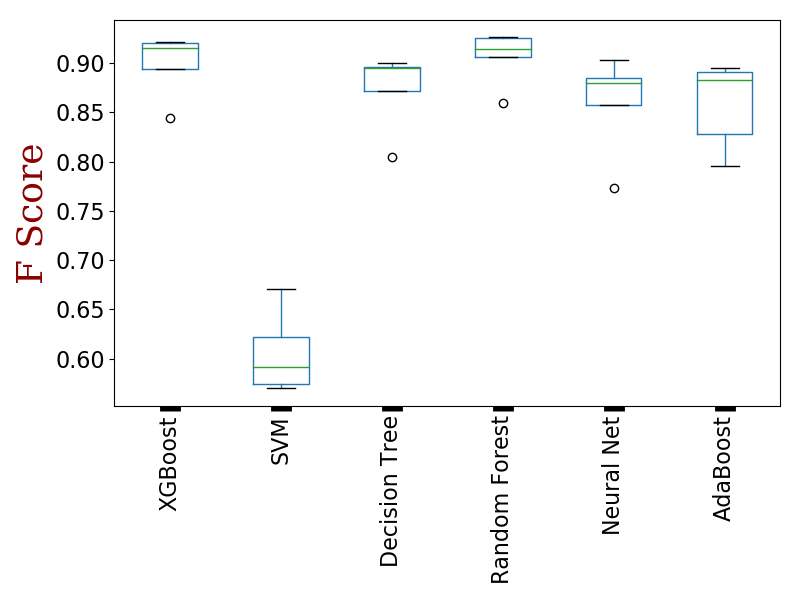}
 \end{tabular}
\caption{Among the trained classifiers random forest achieved the highest mean accuracy.}
\label{fig:clfs}
\end{figure}

In this experiment, we investigated among six classifiers, which classifier is the best. 
The experiment settings use to conventional cross-validation and to perform the transportation mode prediction task showed on \cite{dabiri2018inferring}. 
XGBoost, SVM, decision tree, random forest, neural network, and adaboost are six classifiers that have been applied in the reviewed literature \citep{zheng2010understanding,xiao2017identifying,zhu2018transportation,etemad2018predicting}. 
The dataset is filtered based on labels that have been applied in \cite{dabiri2018inferring} (e.g.,  walking, train, bus, bike, driving) and no noise removal method was applied. 
The classifiers mentioned above were trained, and the accuracy metric was calculated using random cross-validation similar to \cite{liu2017end}, \cite{xiao2017identifying}, and \cite{dabiri2018inferring}.
The results of cross validation, presented in Figure \ref{fig:clfs}, show that the random forest performs better than other models ($\mu_{accuracy} = 90.4\%$). 
The second best model was XGBoost ($\mu_{accuracy} = 90.00\%$).
A Wilcoxon Signed-Ranks Test indicated that the random forest classifier results were not statistically significantly higher than the XGBoost classifier results. 
Wilcoxon Signed-Ranks Tests indicated that the random forest classifier results were statistically significantly higher than the SVM, Neural Network, and Adaboost classifiers results. %,[(Z=-2.6111, p=0.0090),(Z=-1.9844,p= 0.0472),(Z=-1.9844, p= 0.0472)]. 
Moreover, a Wilcoxon Signed-Ranks Test indicated that the random forest classifier results were not statistically significantly higher than the Decision Tree classifier results.%, Z=-1.7755 , p=0.0758.

\subsection{Feature selection using wrapper and information theoretical methods}
\label{sec:wrapper}

\begin{figure}[h]
\begin{tabular}{ll}
\includegraphics[scale=0.21]{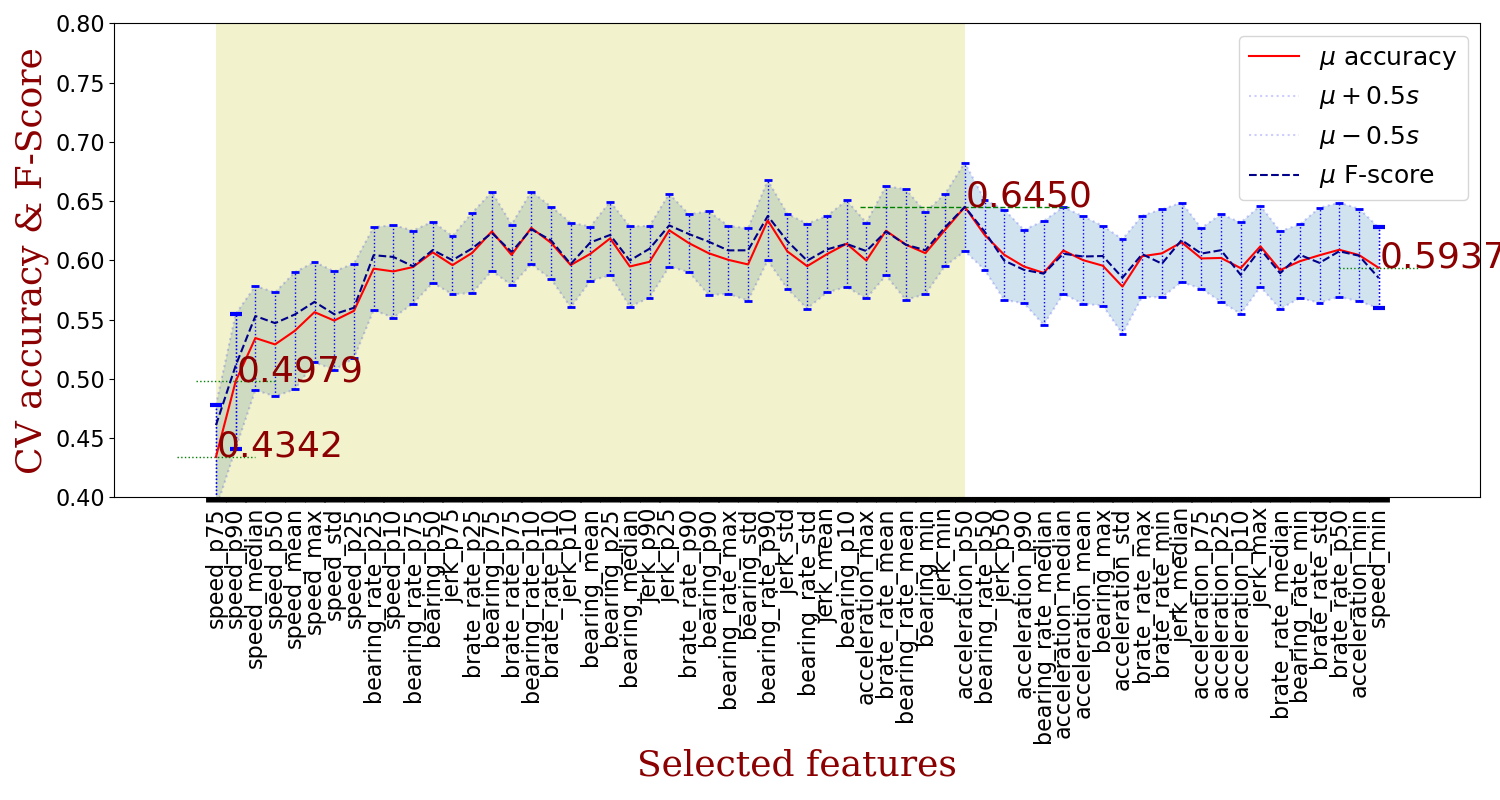}
&
\includegraphics[scale=0.21]{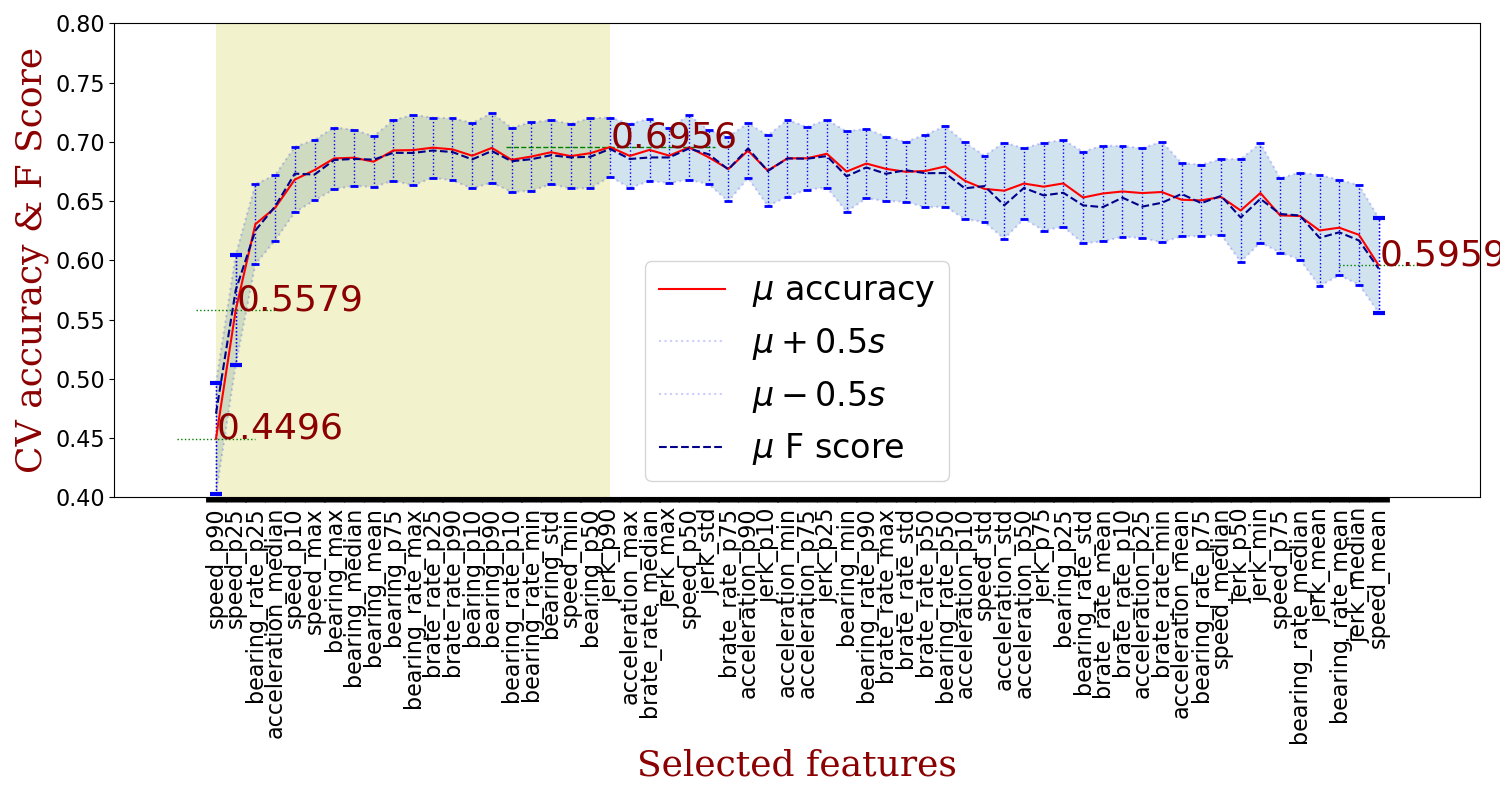}

\end{tabular}
\caption{(a) Accuracy of random forest classifier for incremental appending features ranked by Random Forest feature importance. 
(b) Accuracy of random forest classifier for incremental appending best features}
\label{fig:fs}
\end{figure}

The second experiment aims to select the best features for transportation modes prediction task.% using the Geolife dataset. 
We selected the wrapper feature selection method because it can be used with any classifier. 
Using this approach, we first defined an empty set for selected features. 
Then, we searched all the trajectory features one by one to find the best feature to append to the selected feature set.
The maximum accuracy score was the metric for selecting the best feature to append to selected features. After, we removed the selected feature from the set of features and repeated the search for union of selected features and next candidate feature in the feature set. 
We selected the labels applied in \cite{endo2016deep} and the same cross-validation technique. 
The results are shown in Figure \ref{fig:fs} (a).
The results of this method suggest that the top 20 features get the highest accuracy. 
Therefore, we selected this subset as the best subset for classification purposes using the Random Forest algorithm.

Information theoretical feature selection is one of the methods widely used to select essential features. Random Forest is a classifier that has embedded feature selection using information theoretical metrics.
We calculated the feature importance using Random Forest. 
Then, each feature is appended to the selected feature set and calculating the accuracy score for random forest classifier. The user-oriented cross-validation was used here, and the target labels are similar to \cite{endo2016deep}. 
Figure \ref{fig:fs} shows the results of cross-validation for appending features with respect to the importance rank suggested by the Random Forest.

\subsection{Comparison with \cite{endo2016deep} and \cite{dabiri2018inferring}}

In this third experiment, we filtered transportation modes which have been used by \cite{endo2016deep} for evaluation. 
We divided the training and test dataset in a way that each user can appear only either in the training or test set. 
The top 20 features were selected to be used in this experiment which is the best features subset mentioned in section \ref{sec:wrapper}. 
Therefore, we approximately divided 80\% of the data as training and 20\% of the data as the test set. 
Thus, we compare our accuracy per segment results against \cite{endo2016deep} mean accuracy, 67.9\%.
A one-sample Wilcoxon Signed-ranks test indicated that our accuracy results (69.50\%) are higher than  \cite{endo2016deep}'s results (67.9\%), p=0.0431.

The label set for \cite{dabiri2018inferring}'s research is walking, train, bus, bike, taxi, subway, and car so that the taxi and car are merged and called driving. 
Moreover, subway and train merged and called the train class. 
We filtered the Geolife data to get the same subsets as \cite{dabiri2018inferring} reported based on that. Then, we randomly selected 80\% of the data as the training and the rest as test set- we applied five-fold cross-validation. 
The best subset of features was applied the same as the previous experiment.
Running the random forest classifier with 50 estimators, using SKlearn implementation \cite{scikit-learn}, gives a mean accuracy of 88.5\% for the five-fold cross-validation. 
A one-sample Wilcoxon Signed-ranks test indicated that our accuracy results (88.50\%) are higher than  \cite{dabiri2018inferring}'s results (84.8\%), p=0.0796. 

We avoided using the noise removal method in the above experiment because we believe we do not have access to labels of the test dataset and using this method only increases our accuracy unrealistically. 

\subsection{Effects of types of cross-validation}

\begin{figure}[h]
\centering
 \begin{tabular}{ll}
\includegraphics[scale=0.30]{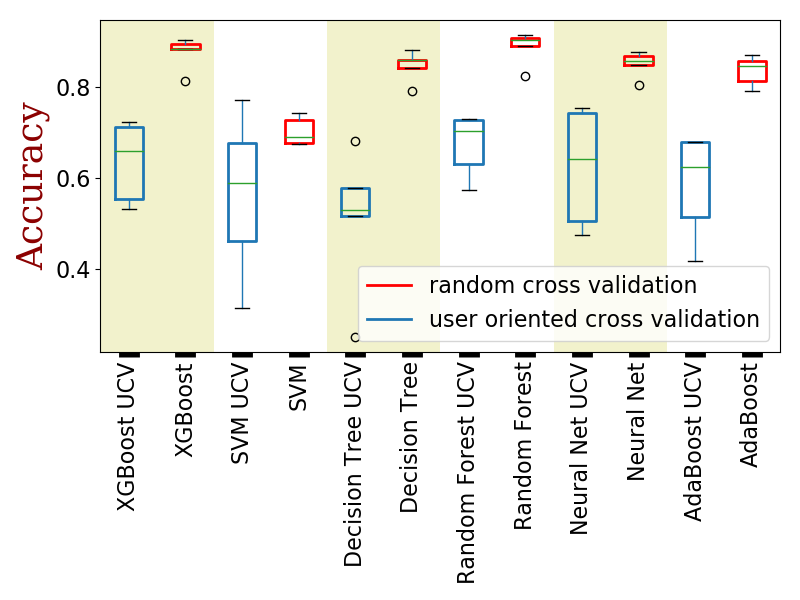}
 &
 \includegraphics[scale=0.30]{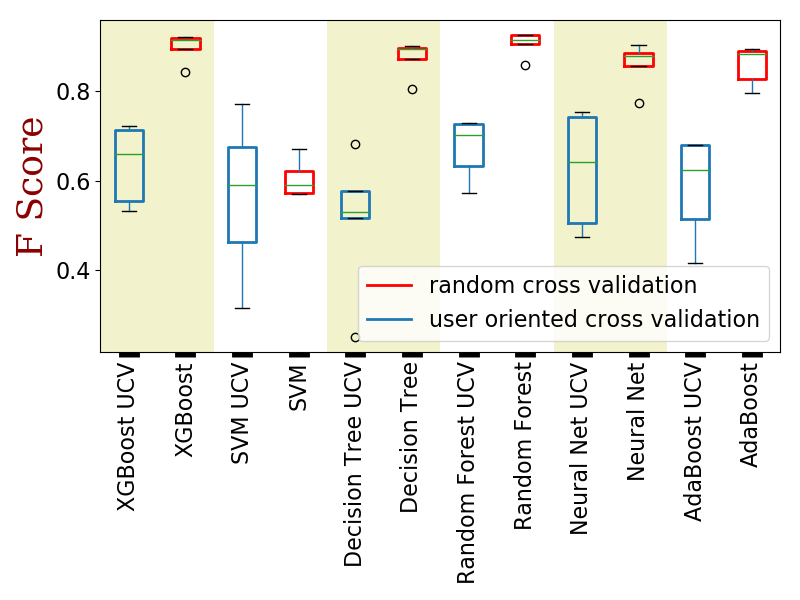}
 \end{tabular}
\caption{The different cross validation results for \textit{user oriented cross-validation} and \textit{random cross-validation}}
\label{fig:crossval}
\end{figure}

To visualize the effect of type of cross-validation on transportation modes prediction task, we set up a controlled experiment. We use the same classifiers and same features to calculate the cross-validation accuracy. Only the type of cross-validation is different in this experiment, one is random, and another is user-oriented cross-validation. 
Figure \ref{fig:crossval} shows that there is a considerable difference between the cross-validation results of user-oriented cross-validation and random cross-validation. 
The result indicates that random cross-validation provides optimistic accuracy and f-score results. 
Since the correlation between user-oriented cross-validation results is less than random cross-validation, proposing a specific cross-validation method for evaluating the transportation mode prediction is a topic that needs attention.

\section{Conclusions}
\label{sec:conclusion}

In this work, we reviewed some recent transportation modes prediction methods and feature selection methods. 
The framework proposed in \cite{etemad2018predicting} for transportation modes prediction was extended, and five experiments were conducted to cover different aspects of transportation modes prediction. 

First, the performance of six recently used classifiers for the transportation modes prediction was evaluated. 
The results show that the random forest classifier performs the best among all the evaluated classifiers.
The SVM was the worst classifier, and the accuracy result of XGBoost was competitive with the random forest classifier.
In the second experiment, the effect of features using two different approaches, the wrapper method and information theoretical method were evaluated.
The wrapper method shows that we can achieve the highest accuracy using the top 20 features. 
Both approaches suggest that the  $F_{p90}^{speed}$ (the percentile 90 of the speed as defined in section \ref{sec:not}) is the most essential feature among all 70 introduced features. 
This feature is robust to noise since the outlier values do not contribute to the calculation of percentile 90.
In the third experiment, the best model was compared with the results showed in  \cite{endo2016deep} and \cite{dabiri2018inferring}. 
The results show that our suggested model achieved a higher accuracy. 
Our applied features are readable and interpretable in comparison to \cite{endo2016deep} and our model has less computational cost.
%Finally, the best model is compared with \cite{dabiri2018inferring} in the fourth experiment. 
%The accuracy results show that our model achieved higher accuracy. 
%The effect of the ground truth noise removal method was investigated. 
%The cleaned dataset achieved a higher accuracy. 
%However, this achievement is optimistic since we do not have access to the test set labels in the pre-processing step. 
Finally, we investigate the effects of user-oriented cross-validation and random cross-validation in the fourth experiments. 
The results showed that  random cross-validation provides optimistic results in terms of the analyzed performance measures.

We intend to extend this work in many directions. 
The spatiotemporal characteristic of trajectory data is not taken into account in most of the works from literature.
We intend to deeply investigate the effects of cross-validation and other strategies like holdout in trajectory data.  
Finally, space and time dependencies can also be explored to tailor features for transportation means prediction.

\bibliography{ref}
\bibliographystyle{plain}
%elsarticle-harv
\end{document}